\documentclass[pdflatex,sn-mathphys-num]{sn-jnl}

\usepackage{graphicx}%
\usepackage{multirow}%
\usepackage{amsmath,amssymb,amsfonts}%
\usepackage{amsthm}%
\usepackage{mathrsfs}%
\usepackage[title]{appendix}%
\usepackage{xcolor}%
\usepackage{xurl}
\usepackage{textcomp}%
\usepackage{manyfoot}%
\usepackage{booktabs}%
\usepackage{algorithm}%
\usepackage{algorithmicx}%
\usepackage{algpseudocode}%
\usepackage{listings}%
\usepackage{multirow}
\usepackage{multicol}
\usepackage{booktabs}
\usepackage{colortbl}
\usepackage{array}
\usepackage{booktabs}
\usepackage{float}
\usepackage{comment}
\usepackage{lineno}
\usepackage{epsfig}

\theoremstyle{thmstyleone}%
\theoremstyle{thmstyletwo}%

\theoremstyle{thmstylethree}%

\begin{document}

\begin{titlepage} 

	\centering 
	
	\scshape 
	
	\vspace*{0.5\baselineskip} 
	
	
	\rule{\textwidth}{1.6pt}\vspace*{-\baselineskip}\vspace*{2pt} 
	\rule{\textwidth}{0.4pt} 
	
	\vspace{0.5\baselineskip} 
	
	{\LARGE Establishing Rigorous and Cost-effective Clinical Trials for Artificial Intelligence Models} 
	
	\vspace{0.1\baselineskip} 
	
	\rule{\textwidth}{0.4pt}\vspace*{-\baselineskip}\vspace{3.2pt} 
	\rule{\textwidth}{1.6pt} 
	
	\vspace{1\baselineskip} 
	
	
	
	\vspace*{0.5\baselineskip} 
	
	
	 \begin{center}Edited By\end{center}
	
	\vspace{0.5\baselineskip} 
	
	{ \begin{center} Wanling Gao \\ Yunyou Huang \\ Dandan Cui \\ Zhuoming Yu \\  Wenjing Liu \\ Xiaoshuang Liang \\ Jiahui Zhao \\ Jiyue Xie \\ Hao Li \\ Li Ma \\ Ning Ye \\ Yumiao Kang \\ Dingfeng Luo \\ Peng Pan \\ Wei Huang \\ Zhongmou Liu \\ Jizhong Hu \\ Gangyuan Zhao \\ Chongrong Jiang \\ Fan Huang \\ Tianyi Wei \\ Suqin Tang \\ Bingjie Xia \\ Zhifei Zhang \\ Jianfeng Zhan \end{center}}
	
	
	\vspace{0.3\baselineskip} 

	\vfill 
	
	
	\epsfig{file=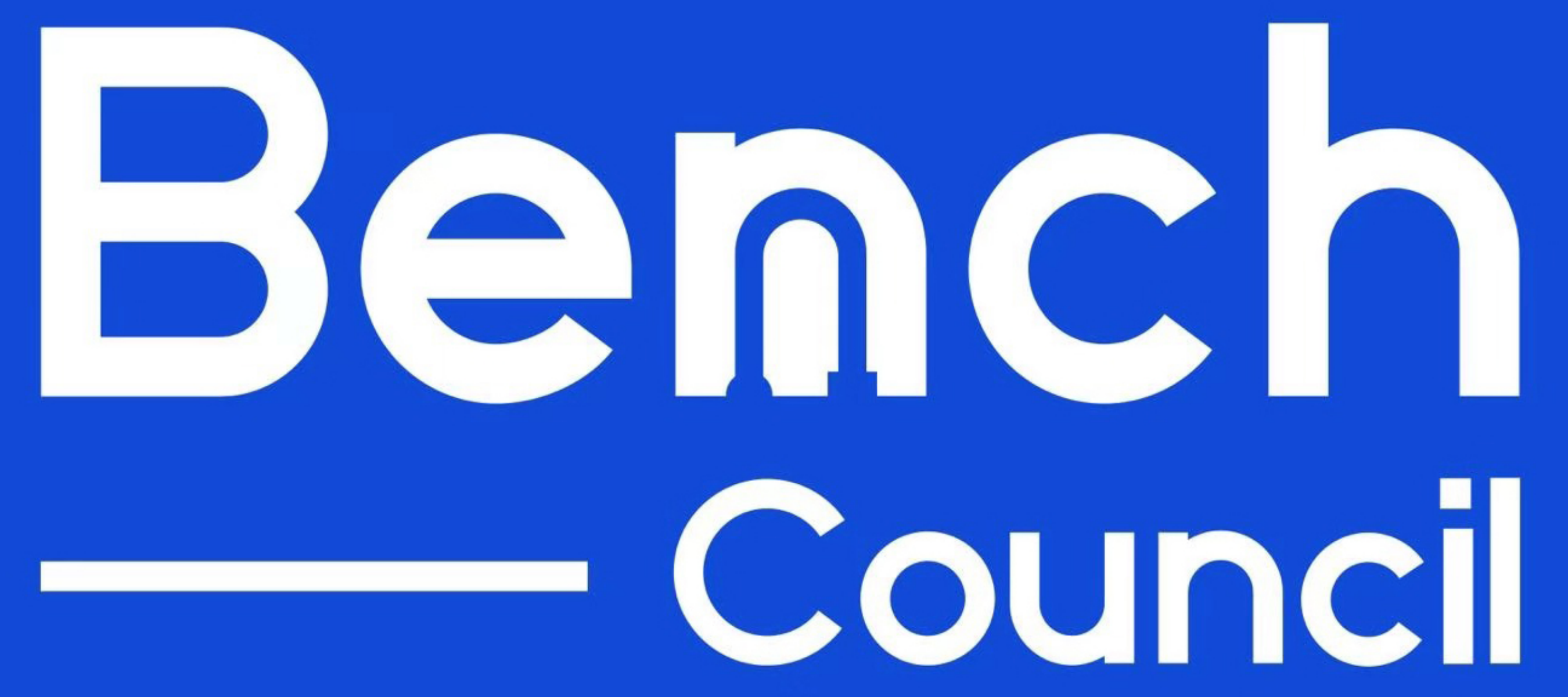,height=2cm}
	\textit{\\BenchCouncil: International Open Benchmark Council\\http://www.benchcouncil.org} 
	\vspace{1\baselineskip} 

	Technical Report No. BenchCouncil-Medicine-Evaluation-2024 
	
	{\large July 11, 2024} 

\end{titlepage}

\title[Article Title]{Establishing Rigorous and Cost-effective Clinical Trials for Artificial Intelligence Models}








\author[1,3,4]{Wanling Gao}
\author[2,18,*]{Yunyou Huang}
\author[1]{Dandan Cui}
\author[2,18]{Zhuoming Yu}
\author[5]{Wenjing Liu}
\author[2,18]{Xiaoshuang Liang}
\author[2,18]{Jiahui Zhao}
\author[2,18]{Jiyue Xie}
\author[2,18]{Hao Li}
\author[5,7]{Li Ma}
\author[6]{Ning Ye}
\author[6]{Yumiao Kang}
\author[8]{Dingfeng Luo}
\author[9]{Peng Pan}
\author[10]{Wei Huang}
\author[11]{Zhongmou Liu}
\author[12]{Jizhong Hu}
\author[13]{Gangyuan Zhao}
\author[14]{Chongrong Jiang}
\author[15]{Fan Huang}
\author[16]{Tianyi Wei}
\author[2]{Suqin Tang}
\author[15,*]{Bingjie Xia}
\author[17,*]{Zhifei Zhang}
\author[1,3,4,*]{Jianfeng Zhan}
\affil[1]{Institute of Computing Technology, Chinese Academy of Sciences, Beijing, 100190, China}
\affil[2]{Guangxi Key Lab of Multi-Source Information Mining and Security, Guangxi
Normal University, Guilin, 541004, China}
\affil[3]{International Open Benchmark Council}
\affil[4]{University of Chinese Academy of Sciences, Beijing, 100086, China}
\affil[5]{Guilin Medical University, Guilin, 541100, China}
\affil[6]{Affiliated Hospital of Guilin Medical University, Guilin, 541000, China}
\affil[7]{XuanJi Technology Co., Ltd., Guilin, 541000, China}
\affil[8]{Xing An County People's Hospital, Guilin, 541300, China}
\affil[9]{Meng Shan County People's Hospital, Wuzhou, 543000, China}
\affil[10]{Guilin People's Hospital, Guilin, 541000, China}
\affil[11]{Yong Fu County People's Hospital, Guilin, 541000, China}
\affil[12]{Ling Chuan County People's Hospital, Guilin, 541000, China}
\affil[13]{Quan Zhou County People's Hospital, Guilin, 541000, China}
\affil[14]{Guan Yang County People's Hospital, Guilin, 541000, China}
\affil[15]{The Second Affiliated Hospital of Guilin Medical University, Guilin, 541000, China}
\affil[16]{International College, Guangxi University, Nanning, 530004, China}
\affil[17]{Capital Medical University, Beijing, 100069, China}
\affil[18]{Key Lab of Education Blockchain and Intelligent Technology, Ministry of
Education, Guangxi Normal University, Guilin, 541004, China}
\affil[*]{corresponding authors: Jianfeng Zhan (zhanjianfeng@ict.ac.cn) and Yunyou Huang (huangyunyou@gxnu.edu.cn) and Zhifei Zhang (zhifeiz@ccmu.edu.cn) and Bingjie Xia (xbj8879@163.com)}




\abstract{
A profound gap persists between artificial intelligence (AI) and clinical practice in medicine, primarily due to the lack of rigorous and cost-effective evaluation methodologies. 
State-of-the-art and state-of-the-practice AI model evaluations 
are limited to laboratory studies on medical datasets or direct clinical trials with no or solely patient-centered controls. Moreover, 
the crucial role of clinicians in collaborating with AI, pivotal for determining its impact on clinical practice, is often overlooked.
For the first time, we emphasize the critical necessity for rigorous and cost-effective evaluation methodologies for AI models in clinical practice, featuring patient/clinician-centered (dual-centered) AI randomized controlled trials (DC-AI RCTs) and virtual clinician-based in-silico trials (VC-MedAI) as an effective proxy for DC-AI RCTs.
Leveraging 7500 diagnosis records from two-step inaugural DC-AI RCTs across 14 medical centers with 125 clinicians, 
our results demonstrate the necessity of DC-AI RCTs and the effectiveness of VC-MedAI. 
Notably, VC-MedAI performs comparably to human clinicians, replicating insights and conclusions from prospective DC-AI RCTs.
We envision DC-AI RCTs and VC-MedAI as pivotal advancements, presenting innovative and transformative evaluation methodologies for AI models in clinical practice, offering a preclinical-like setting mirroring conventional medicine, and reshaping development paradigms in a cost-effective and fast-iterative manner.
Chinese Clinical Trial Registration: \href{https://www.chictr.org.cn/showproj.html?proj=220422}{ChiCTR2400086816}.
}




\maketitle

\section{Introduction}\label{sec1}

Artificial intelligence (AI) holds immense promise in clinical practice~\cite{huang2024pathologist,briganti2020artificial,holmes2004artificial,qian2021prospective} but falls into permissive, time-consuming, and costly evaluations. 
While randomized controlled trials (RCTs) are considered the gold standard for rigorous evaluation~\cite{sibbald1998understanding,akobeng2005understanding}, applying them to AI models in clinical practice poses new challenges.
On the one hand, clinicians remain the primary decision-makers in medicine, viewing AI as a supportive rather than a leading force in shaping patient care~\cite{ye2019psychosocial,yun2021behavioral}. However, traditional trial methodologies, which typically use controls like no intervention, placebo, or sham solely for patients, fail to accommodate the dual focus on both patients and clinicians. This oversight ignores the crucial synergy between AI models and clinicians, hindering the effective integration of AI into clinical practice~\cite{rajpurkar2022ai}, as highlighted by criticisms~\cite{lam2022randomized,angus2020randomized} faced by AI-driven decision support tools, including those officially approved~\cite{abramoff2018pivotal,usfoodadmin}.
Meanwhile, traditional trials are single-blind or double-blind in general, based on feasibility and necessity. 
However, visible (single-blind) or invisible (double-blind) of AI model information like model name to clinicians can induce variations in psychological and behavioral responses. 
Relying on single-blind or double-blind solely is inadequate for fully exploring the synergy between AI and clinicians and evaluating AI software or devices comprehensively. Our results also corroborate this view.

\begin{figure}[htbp]
\centering
\includegraphics[scale=0.42]{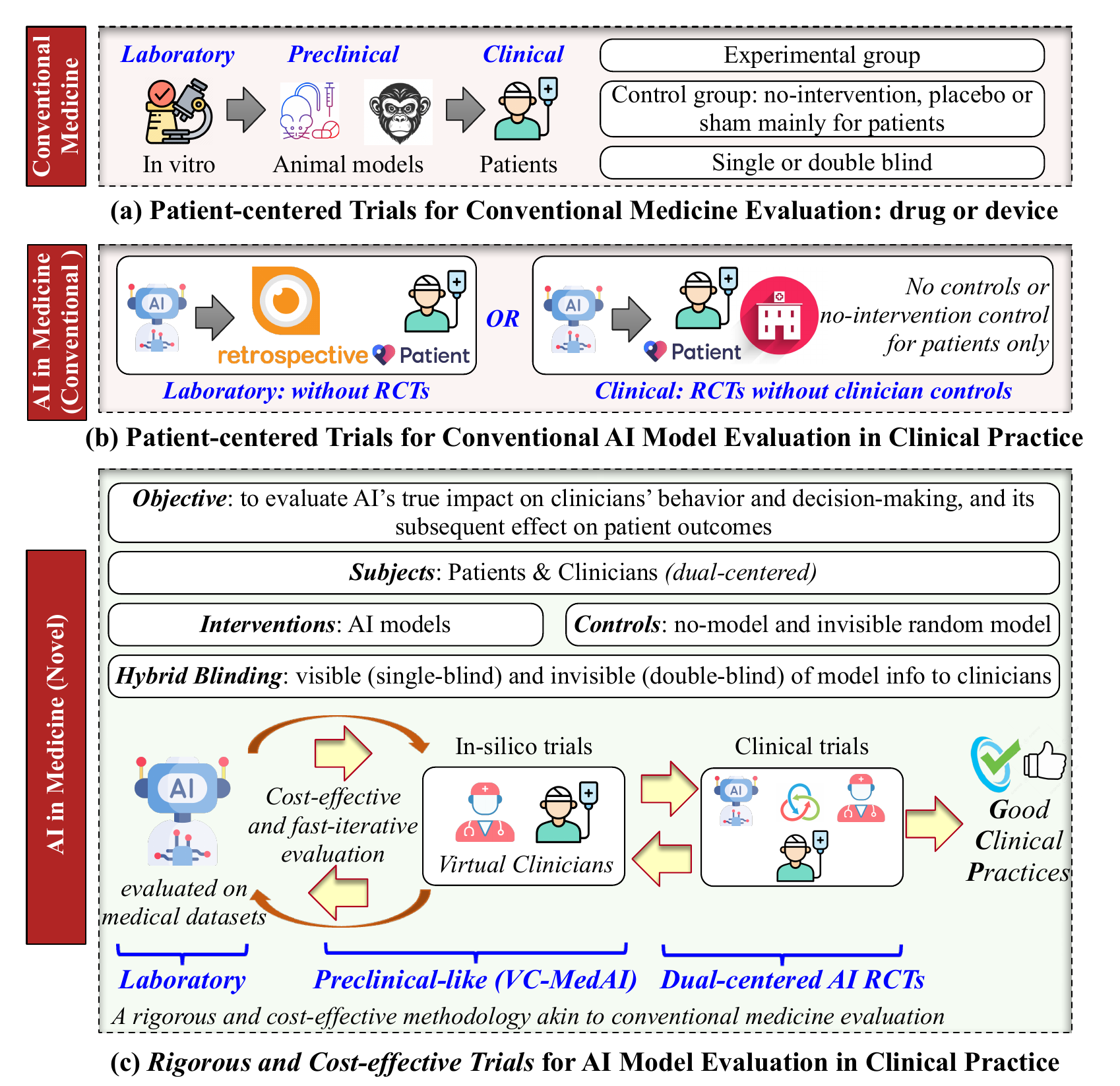}
\caption{\textbf{Transforming Clinical Trials for AI Models in Clinicial Practice.} (a) and (b): patient-centered trials for conventional medicine evaluation and AI model evaluation in clinical practice. 
(c) rigorous and cost-effective trials for AI model evaluation in clinical practice: laboratory, preclinical-like trials using in-silico VC-MedAI, and dual-centered AI randomised controlled trials (DC-AI RCTs).  
}
\label{dpcompare}
\end{figure}

On the other hand, the lack of preclinical trials mandates a swift transition from laboratory evaluations to human clinical trials. This transition carries inherent risks, given the potential for serious clinical outcomes, meanwhile requiring substantial investments in manpower, resources, finances, and time~\cite{martin2017much,rajpurkar2022ai,wiens2019no}.
Consequently, conducting RCTs and publishing reports experience substantial delays~\cite{battelino2023guideline}, often rendered obsolete by rapid technological advancements~\cite{battelino2023guideline}. 
Meanwhile, this prolonged process impedes AI model development from advancing collaboration with clinicians and enhancing clinical outcomes, diverging from the fast-iterative approach pivotal in software development. Moreover, 
the persistent need for RCTs with each software iteration perpetuates delays and exacerbates these challenges.

To tackle the above challenges, for the first time, we emphasize the critical necessity for rigorous and cost-effective evaluation methodologies for AI models in clinical practice, featuring patient/clinician-centered
(dual-centered) AI randomized controlled trials (DC-AI RCTs) and virtual
clinician-based in-silico trials (VC-MedAI), shown in Fig.~\ref{dpcompare}. 
Among them, DC-AI RCTs incorporate well-designed interventions, hybrid single and double blinding, and patient/clinician-centered controls using no-model and random-model.
Experimental groups for each AI model are categorized based on visibility of AI model information to clinicians: visible and invisible. Control groups include no-model and invisible random-model (placebo or sham effect) categories, as visible random-model holds no significance for clinicians. Clinicians and patients are randomly assigned to receive either model-assisted diagnosis or standard care.
Beginning with sepsis, a leading cause of morbidity and mortality~\cite{mayr2014epidemiology,rudd2020global}, we conduct the first rigorous and comprehensive two-step DC-AI RCTs prospectively, across 14 medical centers involving 125 clinicians. 
Step \#1 adopted three AI models and two controls across 2800 patients in 8 clinical settings. 
The 6000 diagnosis records from Step \#1 formed the basis for VC-MedAI. 
In Step \#2, a new AI model assisted with 1500 diagnosis records to prospectively evaluate VC-MedAI.

\begin{figure}[htbp] 
\centering
\includegraphics[scale=0.35]{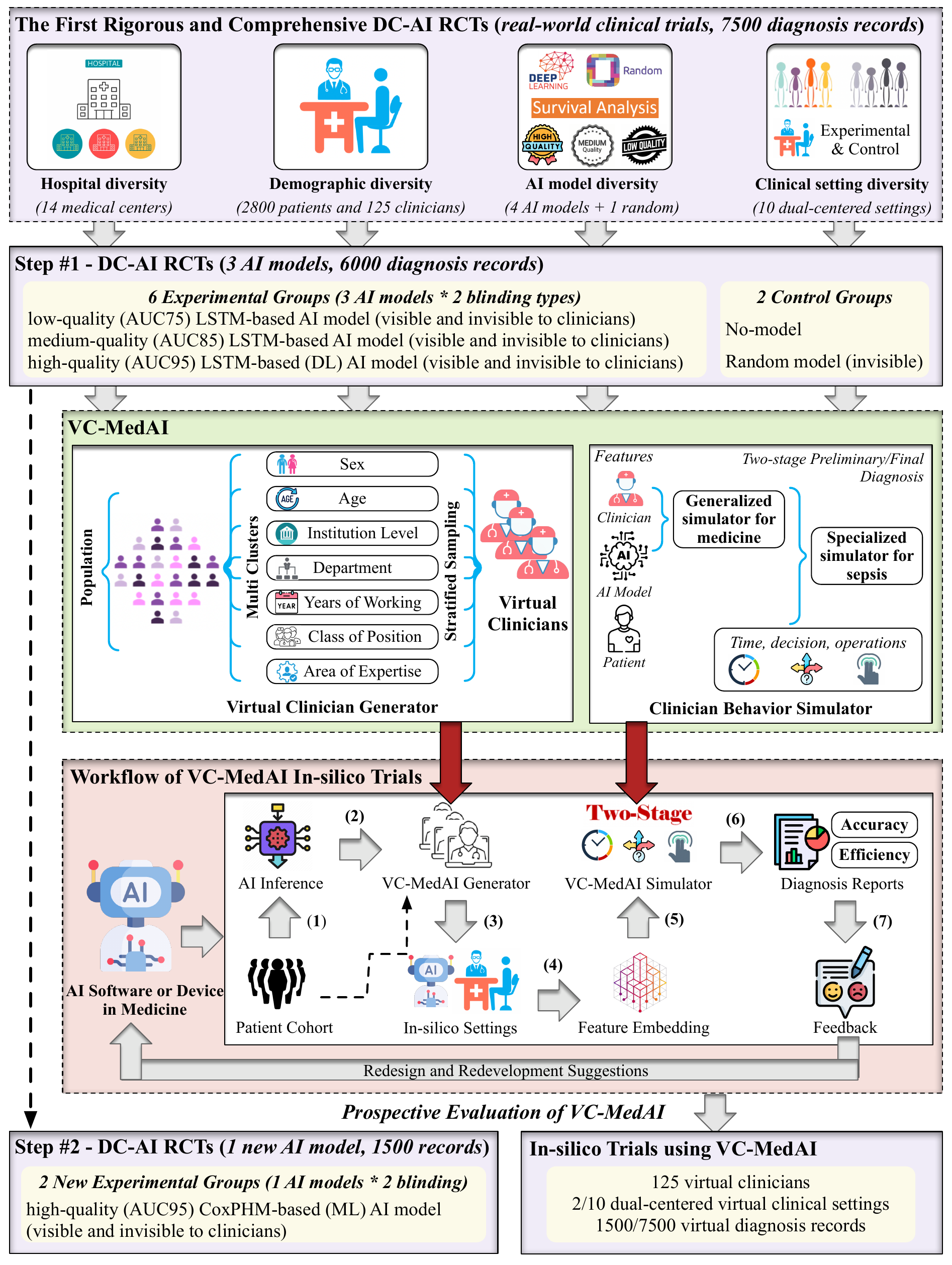}
\caption{\textbf{Workflow of VC-MedAI In-silico Trials.} 
VC-MedAI is constructed and modeled from Step \#1 DC-AI RCTs, containing virtual clinician generator and clinician behavior simulator. We evaluate the effectiveness of VC-MedAI prospectively compared to Step \#2 DC-AI RCTs.
} 
\label{insilico-flow}
\end{figure}

VC-MedAI not only fills gaps left by traditional approaches like preclinical but also complements in-silico trials, known for their effectiveness and cost-efficiency in medicine~\cite{sarrami2021silico,viceconti2016silico,pappalardo2019silico,clermont2004silico}.
For specificity and generality, VC-MedAI provides specialized in-silico trials for sepsis and generalized in-silico trials for other diseases. 
Given a new AI software or device in clinical practice, the workflow of VC-MedAI in-silico trials is illustrated in Fig.~\ref{insilico-flow}.

Our results reveal the necessity of DC-AI RCTs and the effectiveness of VC-MedAI: 
(1) Even an invisible random model can improve clinicians' diagnostic accuracy with 3.37\% AUC (area under the curve) improvement. However, the values for evaluated AI models with two blinding types range from 1.95\% to 10.9\%. This suggests that an AI model's influence on clinical outcomes may not surpass that of a random model. 
Therefore, DC-AI RCTs are crucial to accurately reflect the clinical significance of AI models.
(2) VC-MedAI supports generating a set of virtual clinicians with demographic representativeness, modeling the diagnosis behaviors of human clinicians encountering different interventions and controls, and discovering consistent outcomes and conclusions with prospective clinical trials.  
DC-AI RCTs and VC-MedAI reshape the evaluation methodology of AI models in clinical practice in a rigorous and cost-effective manner. To the best of our knowledge, this is the first attempt to explore such a study and we anticipate it to be a good starting point.

\section{Results}\label{sec2}

\textbf{Diagnosis data from DC-AI RCTs.} 
We perform the first rigorous and comprehensive DC-AI RCTs on sepsis. We set two-step trials and a series of comparative clinical settings, covering a large real-world population~\cite{zhan2024evaluatology}, for example, eight experimental groups with 4 AI models and 2 blinding types, and two control groups with no-model and random model. 
The details are illustrated in Methods section.

\vspace{0.3in}

\noindent \textbf{The necessity of clinician-centered and DC-AI RCTs.}
Our Step \#1 DC-AI RCTs show that 
(1) when diagnosed with an invisible random model, clinicians may show improved diagnosis accuracy on average, with 3.37\% AUC 
improvement from 64.45\% to 67.82\%, suggesting their cautious approach when dealing with models, particularly in cases of inconsistent decision-making.
(2) when diagnosed with the assistance of an AI model, 
clinicians' diagnosis accuracy achieves AUC improvement varying from 1.95\% to 10.9\%,
indicating that the AI model's impact on clinical outcomes might not surpass that of a random model (3.37\%).  
(3) clinicians achieve optimal diagnosis accuracy, sensitivity, and specificity with a visible medium-quality AI model (i.e., 69\%. 77\%, and 61\%, respectively) rather than a visible high-quality one (i.e., 66\%, 73\%, and 59\%) on average, underscoring the critical importance of clinician interactions over model quality alone.
(4) the influence of an AI model on decision-making can vary significantly among different individual or group clinicians due to varying characteristics of clinicians, patients, and the AI model itself, as well as psychological factors.
For example, the difference even achieves to 2X+ on average in terms of diagnosis accuracy and model acceptance rate, which is significant given the value range of 0 to 1. 
Hence, the intricate interactions between AI models and clinicians will incur uncertain and significant implications for patients' treatment strategies, timeliness, and survival rates~\cite{knight2002safety,zhan2024evaluatology}. DC-AI RCTs with rigorous experimental and control groups are essential to ensure fair, standardized, and safe assessment of AI software and devices. 
Details are described in a companion paper about the analysis of DC-AI RCTs on sepsis. 


\vspace{0.3in}

\begin{figure}[H]
\centering
\includegraphics[scale=0.423]{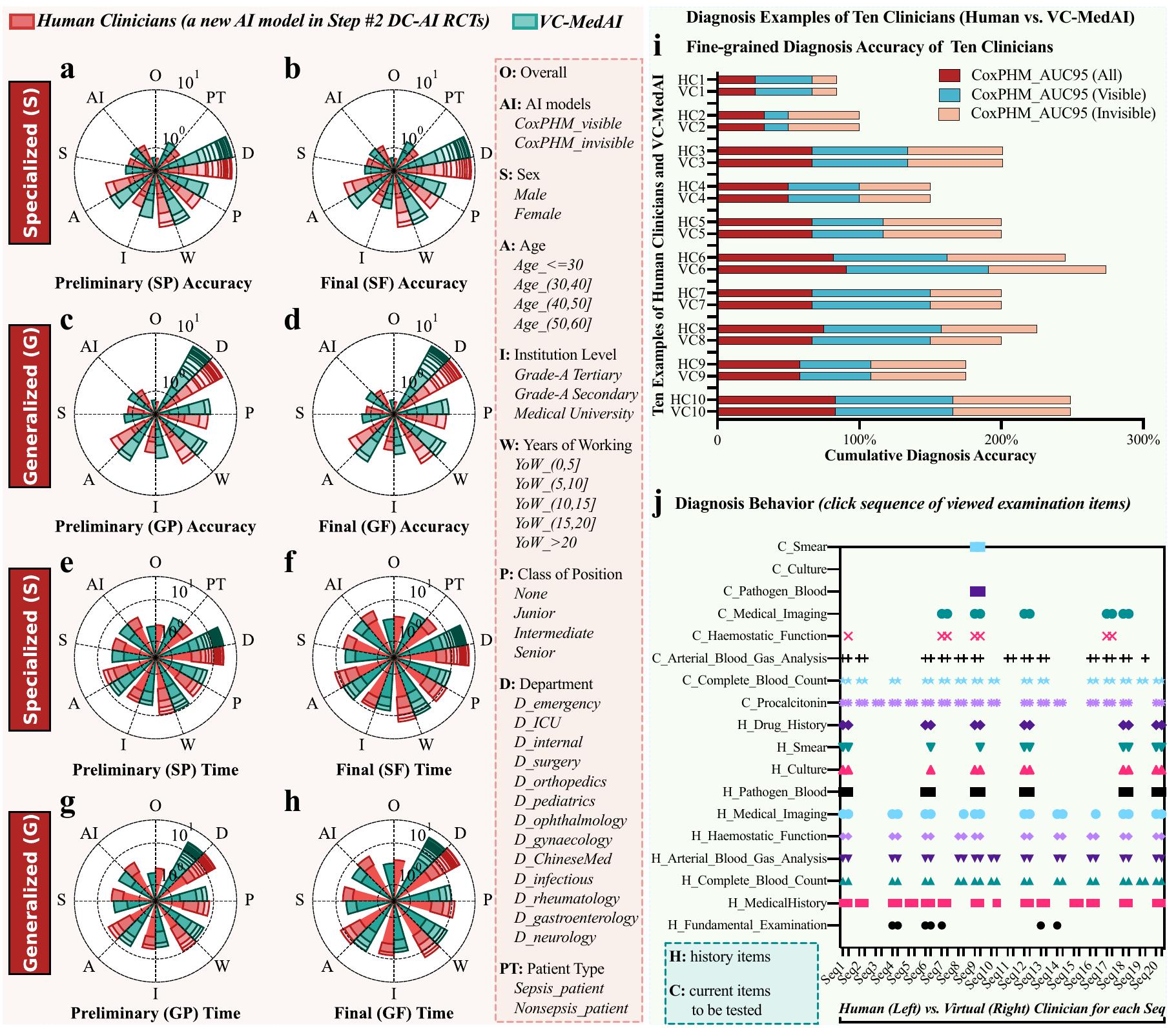}
\caption{\textbf{VC-MedAI Behaves Similar with Human Clinicians Compared to  Prospective Step \#2 DC-AI RCTs.} (a) and (b): the averaged preliminary and final (two-stage) diagnosis accuracy comparisons between VC-MedAI specialized simulator and clinicians. 
(c) and (d): two-stage accuracy comparisons between VC-MedAI generalized simulator and clinicians. 
(e) and (f): two-stage time comparisons between VC-MedAI specialized simulator and clinicians. 
(g) and (h): two-stage time comparisons between VC-MedAI generalized simulator and clinicians. 
The breakdown comparisons of O (Overall), AI (AI models), S (Sex), A (Age), I (Institution), W (Years of Working), P (Class of Position), D (Department), and PT (Patient Type) are arranged from inside to outside within each corresponding sector as shown in the legend. Generalized simulator has no PT sector for general simulation. (i) and (j) are specific diagnosis examples of ten clinicians and corresponding VC-MedAI with the same features. (i): diagnosis accuracy. (j): diagnosis behavior (click sequence of examination items). Each sequence from Seq1 to Seq20 contains the operations of a human clinician (left) and corresponding virtual one (right). Note that history fundamental examination items include body temperature, systolic blood pressure, diastolic blood pressure, heart rate, respiratory rate, consciousness level, and qSOFA (quick Sequential Organ Failure Assessment).
}
\label{vd-test}
\end{figure}

\noindent \textbf{VC-MedAI behaves similar with human clinicians in terms of preliminary and final diagnosis decision.} 
The preliminary and final diagnosis decisions of VC-MedAI specialized simulator and human clinicians with the assistance of CoxPHM  
are compared in Fig.~\ref{vd-test}(a) and Fig.~\ref{vd-test}(b), respectively.
The simulator demonstrates high similarity with human clinicians with an AUC of 0.81 (95\% CI: 0.78-0.83) for preliminary diagnosis and 0.82 (95\% CI: 0.8-0.84) for final diagnosis 
compared to prospective Step \#2 DC-AI RCTs.
Overall, their diagnosis accuracies 
on average are quite similar with a deviation of 3\% for both preliminary and final diagnosis, as shown in the ``O" sector of Fig.~\ref{vd-test}(a) and Fig.~\ref{vd-test}(b). 
This implies that if a new AI algorithm performs well in VC-MedAI specialized simulator, it will likely show similar performance in realistic clinical settings when collaborating with human clinicians.

We further perform detailed comparisons from the perspectives of AI models (AI), clinician properties like age (A), 
and patient types (PT), covering a total of 35 dimensions as shown in the legend of Fig.~\ref{vd-test}. The comparison deviation averages 6.5\%  for preliminary diagnosis and 7.4\% for final diagnosis. Specifically, we found that in 32 out of 35 dimensions for preliminary diagnosis, the comparison deviation is around 10\%, with 
23 dimensions falling within a 0\% to 6\% range.  
Note that the other three dimensions with higher comparison deviations are orthopedics (33\%), pediatrics (16\%), medical university (25\%), since  
the clinician samples from these dimensions are either small in number or have specific peculiarities, e.g., data collection from minors is restricted and protected by law. 
The situation of final diagnosis remains quite similar.
This implies that at a fine-grained level, VC-MedAI specialized simulator also reflects the behavior of human clinicians in realistic clinical settings.

The VD-MedAI generalized simulator achieves an AUC of 0.79 (95\% CI: 0.77-0.81) for preliminary diagnosis and 0.82 (95\% CI: 0.79-0.84) for final stage 
on CoxPHM-assisted evaluations, compared to real-world Step \#2 trials. 
Due to the reduction in patient features, the generalized simulator exhibits a minor decrease in model quality 
compared to the specialized one. Nevertheless, it remains within practical usability and supports the human-AI interaction simulation through in-silico trials for various diseases in medicine.
Fig.~\ref{vd-test}(c) and Fig.~\ref{vd-test}(d) shows the diagnosis accuracy comparisons.
The comparison deviation averages 6.6\% for preliminary diagnosis and 10.38\% for final diagnosis, from overall dimension and 33 detailed dimensions. 
This implies that not only for sepsis, but also for other diseases, if the corresponding AI algorithm performs well in VC-MedAI generalized simulator, it will similarly demonstrate comparable performance in realistic clinical settings when interacting with human clinicians.

VC-MedAI demonstrates a high degree of similarity to human clinicians not only on average level but also at the individual level. 
Specifically, Fig.~\ref{vd-test}(i) shows a comparison of fine-grained preliminary or final diagnosis accuracy of ten human clinicians and VC-MedAI, with the assistance of CoxPHM-based AI models under different blinding types. Note that HD represents a human clinician and VD represents corresponding virtual clinician who has similar features. 
We find that 
the model's impact varies among different clinicians; for some, visible models provide more significant assistance (e.g., Case\#1), while for others, invisible models are more effective (e.g., Case\#2).
VC-MedAI also reflects similar characteristics. The situation remains the same for all ten compared clinicians.
This implies that even for individual diagnostic case by a single clinician, VC-MedAI maintains consistency with realistic diagnostic behaviors. 



\vspace{0.3in}

\noindent \textbf{VC-MedAI behaves similar with human clinicians in terms of operation behaviors.} In addition to the diagnosis decision and results, the intermediate data and behaviors during the diagnosis process are also important features. For example, the examination data of patients serve as crucial or decisive references for clinician's decision-making. Hence, the operation behaviors involve a sequence of viewed or checked examination items. VC-MedAI also supports the operation behavior simulation of human clinicians, compared to prospective Step \#2 DC-AI RCTs. 
The recall-oriented understudy for gisting evaluation (ROUGE) score is \{`r': 0.80, `p': 0.69, `f': 0.74\} for ROUGE-1 and \{`r': 0.77, `p': 0.66, `f': 0.71\} for ROUGE-L. 
Fig.~\ref{vd-test}(j) shows detailed comparisons of twenty example sequences.  
We find that their operation sequences of human clinicians and VC-MedAI specialized simulator achieves high degree of similarity from perspectives of not only specific examination items but also sequence length.
Overall, ten out of twenty sequences are exactly the same and the remaining ones only have slightly deviation which either checked one or two fewer items or one or two more items.
Meanwhile, the averaged lengths of operation sequence for human clinicians and VC-MedAI specialized simulator corroborate this point, which are 5.07 and 6.51, respectively. 
This implies that for a given AI algorithm assisting with medical diagnosis, the examination items viewed by VC-MedAI will also be similarly viewed in realistic clinical settings.

Considering examination item variations among different diseases,
VC-MedAI generalized simulator simulates the 
ratio of advanced items need to be further tested, e.g., 
haemostatic function, 
instead of specific sequence for generality. 
The AUC of VC-MedAI generalized simulator for the operation behavior simulation achieves 0.85 (95\% CI: 0.83-0.87). 
The comparison deviation averages 2.27\% from 33 dimensions of AI models and clinician features. Please see Appendix Tables B1-B2 for details.
This implies for other diseases not merely sepsis, the item ratio ordered by human clinicians in realistic clinical settings will be similar with the output of VC-MedAI.

\vspace{0.3in}

\noindent \textbf{VC-MedAI behaves similar with human clinicians in terms of preliminary and final diagnosis time.} Diagnosis time is another crucial behavioral characteristic and key factor, especially in emergency situations where quick diagnosis can significantly increase survival rates with even a slight time advantage. 
Fig.~\ref{vd-test}(e) and Fig.~\ref{vd-test}(f) show the diagnosis time comparison of human clinicians and VC-MedAI specialized simulator in terms of preliminary and final diagnosis, respectively, compared to Step \#2 DC-AI RCTs using CoxPHM.
Note that 
the diagnosis time only considers the clinician's diagnostic decision-making based on examination data and does not account for the waiting time required to obtain these data. 
Considering the similarity of human clinicians and VC-MedAI in operation behaviors illustrated above, the whole time 
plus waiting time required by additional examination items likewise maintains similarity. 
The time units 
are in minutes, displayed on a logarithmic scale. 
Considering the potential time variability when a person repeats an operation multiple times, we define ±20\% of clinician's diagnosis time 
as true time range.
In terms of the preliminary diagnosis of VC-MedAI specialized simulator, the mean absolute error (MAE) is 0.5 minutes (95\% CI: 0.46-0.53), occupying 26.6\% of the average value (1.88 minutes). As for the final diagnosis, the MAE is 0.81 minutes (95\% CI: 0.74-0.89), occupying 26.3\% of the average value (3.08 minutes). 
Fig.~\ref{vd-test}(g) and Fig.~\ref{vd-test}(h) show the performance of VC-MedAI generalized simulator. 
The MAE is 0.44 minutes (95\% CI: 0.41-0.48) for preliminary diagnosis, occupying 23.4\% of the average value, and 0.73 minutes (95\% CI: 0.66-0.80) for final diagnosis, occupying 23.8\% of the average value. 
The experimental data indicates that when using VC-MedAI, the AI model's impact on diagnosis time, whether positive or negative, aligns with realistic clinical outcomes.

\begin{table}[]
\definecolor{mycolor1}{HTML}{FAF4F4}
\definecolor{mycolor2}{HTML}{EFAC8F}
\caption{\textbf{Virtual Clinician Population and Demographics Generated by VC-MedAI.}}
\centering
\footnotesize
\renewcommand\arraystretch{1.2}
\begin{tabular}{|p{0.9in}|p{1.901in}|p{0.7in}|p{0.33in}|p{0.43in}|}
\hline
\rowcolor{mycolor2}
\textbf{Category} & \textbf{Characteristics} & \textbf{No. of Clinicians (\% by unit)} & \textbf{Median} & $\textbf{Mean}\pm \textbf{SD}$ \\ \hline
\multirow{2}{2in}{\textbf{Sex}} & Male & 65 (52\%) & & \\ 
& Female & 60 (48\%)  & &  \\ \hline
\multirow{4}{2in}{\textbf{Age}} & $age \leq 30$ & 35 (28\%) & 28 & $26.2 \pm 3.2$ \\ 
& $30 < age \leq 40$ & 42 (33\%)  & 35 & $35.4 \pm 2.8$ \\ 
& $40 < age \leq 50$ & 36 (29\%)  & 43 & $43.5 \pm 2.0$\\ 
& $50 < age \leq 60$ & 12 (10\%)  & 55 & $54.8 \pm 2.9$ \\ \hline
\multirow{5}{2in}{\textbf{Years of Working}} & (0,5] & 39 (31\%)  & 4 &  $3.4 \pm 1.5$ \\ 
& (5,10] & 21 (17\%)  & 10 & $9.1 \pm 1.5$ \\ 
& (10,15] & 18 (14\%)  & 13 & $13.1 \pm 1.1$ \\ 
& (15,20] & 27 (22\%) & 19 & $18.5 \pm 1.7$ \\ 
& \textgreater 20 & 20 (16\%)  & 30 & $28.6 \pm 5.3$ \\ \hline
\multirow{4}{2in}{\textbf{Class of Position}} & None (During residency training) & 21 (17\%)  & & \\ 
& Junior (Resident physician) & 16 (13\%)  & & \\ 
& Intermediate (Attending physician)  & 42 (33\%)  & & \\ 
& Senior (Chief/Associate chief physician) & 46 (37\%)  & & \\ \hline
\multirow{3}{2in}{\textbf{Institution Level}} & Grade-A Tertiary Hospital & 35 (28\%)  & & \\ 
& Grade-A Secondary Hospital & 88 (70\%)  & & \\ 
& Medical University & 2 (2\%)  & & \\ \hline
\multirow{14}{2in}{\textbf{Department}} & Emergency & 32 (25.6\%)  & & \\ 
& Intensive Care Unit (ICU) & 48 (38.4\%)  & &  \\ 
& Internal Medicine & 14 (11.2\%)  & &  \\ 
& Surgery & 7 (5.6\%)  & & \\ 
& Orthopedics & 1 (0.8\%)  & & \\ 
& Pediatrics & 8 (6.4\%)  & & \\ 
& Ophthalmology & 1 (0.8\%)  & & \\ 
& Gynaecology & 5 (4\%)  & & \\ 
& Traditional Chinese Medicine (TCM) & 6 (4.8\%)  & & \\ 
& Infectious Diseases  & 1 (0.8\%)  & & \\
& Rheumatology and Immunology & 1 (0.8\%)  & & \\ 
& Neurology & 1 (0.8\%)  & & \\ 
\hline 
\end{tabular}
\label{vdoctor-charac-table}
\end{table}

\vspace{0.3in}

\begin{figure}[htbp]
\centering
\includegraphics[scale=0.423]{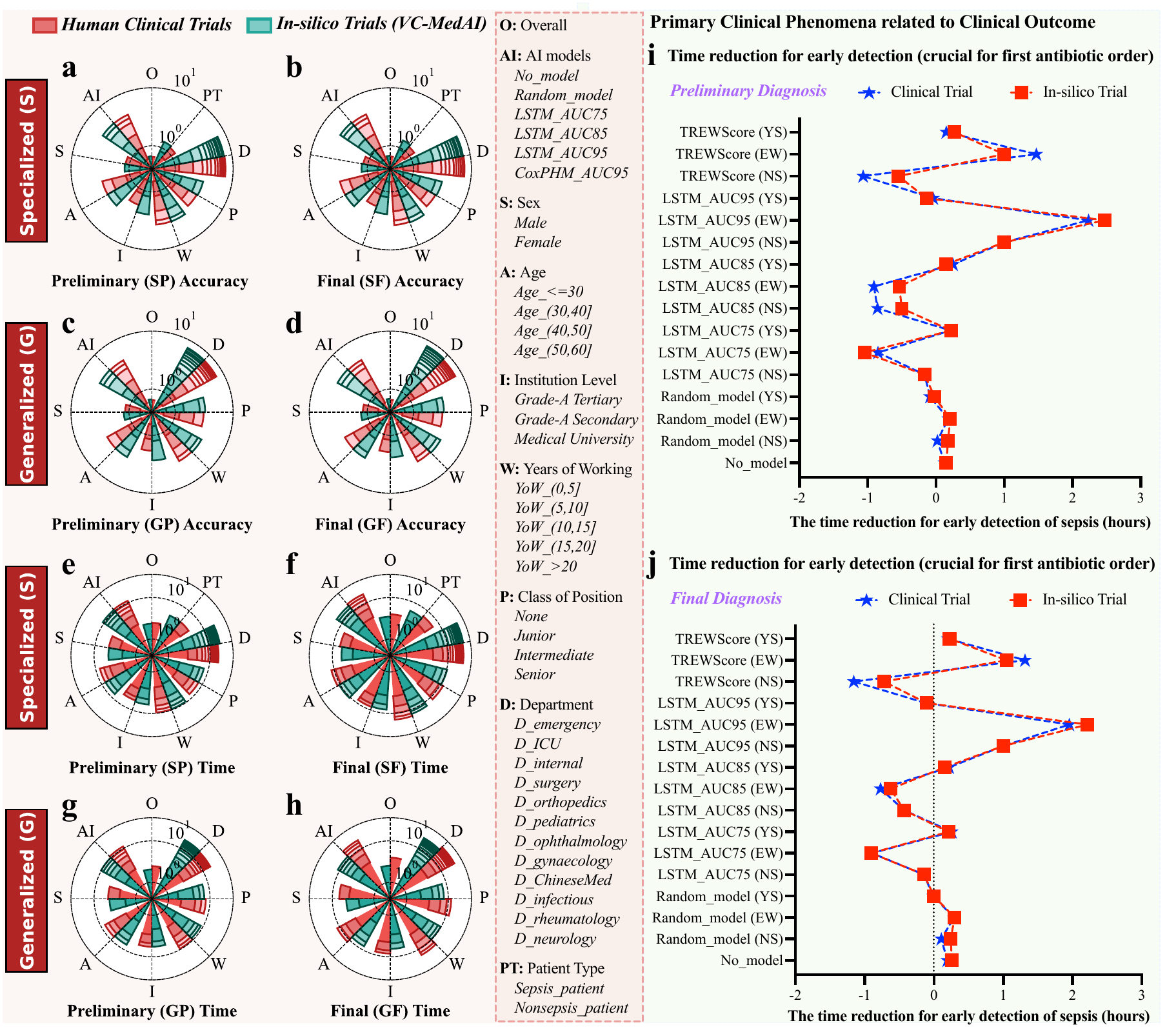}
\caption{\textbf{VC-MedAI supports the discovery of consistent outcomes with real-world clinical trials.} (a) and (b): the averaged preliminary and final diagnosis (two-stage) accuracy comparisons between human clinical and in-silico trials (specialized). (c) and (d):  two-stage diagnosis accuracy comparisons between human clinical and in-silico trials (generalized). (e) and (f): two-stage diagnosis time comparisons between human clinical and in-silico trials (specialized). (g) and (h): two-stage diagnosis time comparisons between human clinical and in-silico trials (generalized). Note that the breakdown comparisons of O, AI, S, A, I, W, P, D, and PT  
are arranged from inside to outside within each corresponding sector. (i) and (j) represent the time reduction for early detection relative to the onset time of sepsis patients, which is a primary clinical phenomenon related to clinical outcome. (i): preliminary and (j): final diagnosis. YS, EW, and NS indicate the model's prediction as sepsis, early warning of sepsis, and non-sepsis, respectively. A larger value signifies larger time reduction and earlier detection of sepsis.
}
\label{vd-outcome}
\end{figure}

\noindent \textbf{VC-MedAI supports the discovery of consistent outcomes with conventional clinical trials.} 
We use VC-MedAI to perform in-silico trials adopting the same settings with the real-world two-step DC-AI RCTs, which are detailedly illustrated in Methods section. Specifically, VC-MedAI generator first generates the same number of 125 virtual clinicians with new features through stratified sampling. 
The generating time is 0.41 seconds.
Table~\ref{vdoctor-charac-table} illustrates the virtual clinician population and demographics. 
VC-MedAI simulator then performs two-stage diagnosis on the same patients using the new generated virtual clinicians, with or without the assistance of different AI or random models. 
The diagnosis records are 7500 in total with the time consumption of 4.83 hours, including 1.67 hours for preliminary and 3.16 hours for final diagnosis. Compared to one iteration of human clinical trials which may span from several months to several years, VC-MedAI achieves at least 150X speedup (4.83 hours vs. 1 month), not to mention longer time for one trial and more trial iterations.
Moreover, the in-silico trials of 7500 diagnosis records using VC-MedAI reflect similar characteristics with the human clinical trials in terms of diagnosis accuracy and diagnosis time, as shown from Fig.~\ref{vd-outcome}(a) to Fig.~\ref{vd-outcome}(h). 
This implies that VC-MedAI can support rapid assessment, validation, and iteration of AI software or device, while maintaining diagnostic behaviors similar to human clinicians in realistic clinical settings.

In realistic clinical settings, mortality is the most critical clinical outcome and is closely related to the timely use of antibiotics. The earlier antibiotics are administered, the higher the patient's survival chances~\cite{henry2022factors,liu2017timing}. Furthermore, studies have shown that the use of antibiotics is closely linked to the early detection of sepsis~\cite{singer2016third,dellinger2013surviving,husabo2020early}. The earlier sepsis is detected, the higher the likelihood of timely antibiotic administration and improved outcomes. 
We compared the time reduction for early detection of sepsis -- a primary clinical phenomena related to the first antibiotic order and patients' outcomes -- between human clinical trials and virtual clinician based in-silico trials.
Specifically, we calculated and averaged the time reduction relative to the onset time of sepsis under different settings, e.g., without model assistance, with the assistance of random, low-quality LSTM-based (AUC 0.75), medium-quality LSTM-based (AUC 0.85), high-quality LSTM-based (AUC 0.95), and high-quality CoxPHM-based (AUC 0.95) AI models. Meanwhile, we differentiated the impact of model predictions on clinical phenomena and used YS, EW, NS to represent the model prediction as sepsis, early warning of sepsis, and non-sepsis, respectively.
The time reduction comparisons of preliminary and final diagnosis are shown in Fig.~\ref{vd-outcome}(i) and Fig.~\ref{vd-outcome}(j), respectively, with the unit in hours. Positive values indicate the sepsis was detected earlier relative to the onset time, while negative values indicate the disease was detected later. The larger the value, the greater the time reduction, indicating earlier detection.
We observe that clinical phenomena in both human and in-silico trials exhibit similar and consistent trends. For instance, the average time reduction achieved using a high-quality LSTM-based AI model is 0.753 hours in human clinical trials and 0.777 hours in in-silico trials compared to diagnoses without model assistance. With a prospective high-quality CoxPHM-based AI model, the reductions are -0.074 hours and -0.077 hours, respectively.
From a trend perspective, we find that the model's performance in in-silico trials is similarly reflected in human clinical trials. For example, we find that for patients who have not yet during the sepsis onset, the model's prediction as an early warning is more readily accepted by clinicians and achieves higher time reduction. This situation remains consistent across human clinical and in-silico trials. 

As a result, virtual clinician based in-silico trials using VC-MedAI can replicate conclusions and findings from realistic human clinical trials, implying that the outcomes of a new AI software or device obtained from in-silico trials would also yield similar discoveries and conclusions in realistic clinical settings.



\vspace{0.3in}

\section{Discussion}

In this study, 
we innovatively emphasize the critical necessity for rigorous and cost-effective evaluation methodologies for AI models in clinical practice, featuring patient/clinician-centered
(dual-centered) AI randomized controlled trials (DC-AI RCTs) and virtual
clinician-based in-silico trials (VC-MedAI).
We perform the first rigorous and comprehensive DC-AI RCTs across 14 medical centers involving 125 clinicians. Step \#1 using three models constitutes the basis for VC-MedAI, further confirming the necessity of a rigorous approach and DC-AI RCTs. 
Step \#2 using a new AI model provides prospective trials for evaluating VC-MedAI, verifying its effectiveness as a substitute for DC-AI RCTs and as a preclinical-like process mirroring conventional medicine. 

Conventional clinical trials mainly regard patient as main subjects and adopt patient-centered trials including experimental or control groups to assess new techniques, device, or drugs~\cite{friedman2015fundamentals,piantadosi2024clinical,phillips2014primate,pouladi2013choosing}. 
In contrast, AI models in clinical practice relies on clinician-in-the-loop mechanism and the patients are no longer the solely subjects. 
Despite numerous proposals for AI software and devices in clinical practice~\cite{henry2022factors,kaji2019attention,idxdr}, none have conducted DC-AI RCTs. 
According to a survey~\cite{lam2022randomized} in 2022, only 39 (0.33\%) out of 11,839 articles have conducted RCTs. Among these, 92\% (36/39) compared AI-assisted tools to controls using standard care, while 5\% (2/39) used a sham treatment without AI assistance as a control~\cite{lam2022randomized}, all overlooking clinician synergy.
In this condition, it is challenging to determine the extent of the AI software's impact and whether it has a positive effect, no matter from the accuracy, diagnosis time, or cost perspective.
Our study addresses this deficiency and offers the first rapid and cost-effective alternative for AI model evaluations in clinical practice.

The limitation of the study is as follows.
VC-MedAI reflects a deviation around 25\% considering average length of operation sequences and diagnosis time, compared to prospective Step \#2 DC-AI RCTs. We will improve its quality through more clinical trials. Notably, these deviations have little impact on the clinical outcomes according to our results.

Clinical outcome is of paramount importance in clinical practice. 
We highlight DC-AI RCTs -- introducing hybrid blinding types and clinician controls -- are crucial for shielding confounding factors and uncovering genuine impacts of AI software or device on clinical outcome. 
For instance, evaluating an AI model solely against a no-model control in hospitals might suggest it enhances diagnosis accuracy, sensitivity, and specificity. However, comparing it with an invisible random model may reveal even a random model can improve clinical outcomes to some extent, whereas the AI model shows minimal, if any, enhancing effect. Thus, the conclusion shifts, indicating the AI model has little to no impact on clinical outcomes. 
Meanwhile, as an effective proxy of DC-AI RCTs, VC-MedAI remains consistency with real-world clinical trials in terms of a primary clinical phenomena -- time reduction for early detection that closely linked to the use of antibiotics -- related to outcomes~\cite{singer2016third,dellinger2013surviving,husabo2020early,henry2022factors,liu2017timing}. 
From a clinical perspective, AI software and device assessments need to center on clinicians and be grounded in DC-AI RCTs, reshaping evaluation methodologies and pioneering innovative directions. 
In terms of AI development, it's crucial that AI models not only enhance the quality and performance but also prioritize interaction with clinicians to bolster their acceptance.
Meanwhile, AI development should account for regional variations and clinician demographics to enhance clinical applicability and accelerate progress in clinical practice. We anticipate our evaluation methodology and tool will accelerate and enhance the assessment of AI effects beneficially.


\section{Methods}\label{sec11}

\textbf{AI models and random control used in DC-AI RCTs.} 
The AI models include state-of-the-art/practice algorithms --  the survival analysis based cox proportional hazards model (CoxPHM) model~\cite{henry2022factors,henry2015targeted} and deep learning based long short term memory (LSTM) model~\cite{kaji2019attention}, specifically, low-quality (AUC 0.75), medium-quality (AUC 0.85), high-quality (AUC 0.95) LSTM-based models, a high-quality (AUC 0.95) CoxPHM-based model, and a random model (AUC 0.5) as control.

\vspace{0.3in}

\noindent \textbf{Randomization and blinding.} 
We conduct DC-AI RCTs across 14 medical centers with 125 clinicians.
The patient cohort is well chosen from MIMIC~\cite{johnson2016mimic,johnson2020mimic,johnson2019mimic,johnson2023mimic} datasets, which comprises 3000 cases divided into five groups. 
First, MIMIC assigns each patient a unique randomized ID. Second, we perform multi-threaded sequential reads using these randomized IDs, leveraging the inherent randomness. Third, we sequentially select and categorize patient IDs that meet specific criteria into five distinct groups, as defined by our collaborating clinicians, which include medical imaging and comprehensive examination data.
In Step \#1, the first group is diagnosed by clinicians without and with the above random model or LSTM-based models, 
having no idea about the model properties. In total, group \#1 contains five settings. For comparison, patient group \#2 to \#4 are diagnosed with the assistance of three LSTM-based AI models and the clinicians are aware of the model properties. In total, Step \#1 contains eight settings. 
In Step \#2, a new CoxPHM-based model is employed to assist the diagnosis of patient group \#1 and group \#5, with the model's properties being invisible to clinicians for group \#1 and visible for group \#5. 
To unveil clinician's behaviors and their intricate interactions with AI, every clinician is assigned partial patient cases randomly within each clinical setting.
A companion paper about data descriptors illustrates the collected records~\cite{gao2024ai}.

\vspace{0.3in}

\begin{figure}[htbp]
\centering
\includegraphics[scale=0.3]{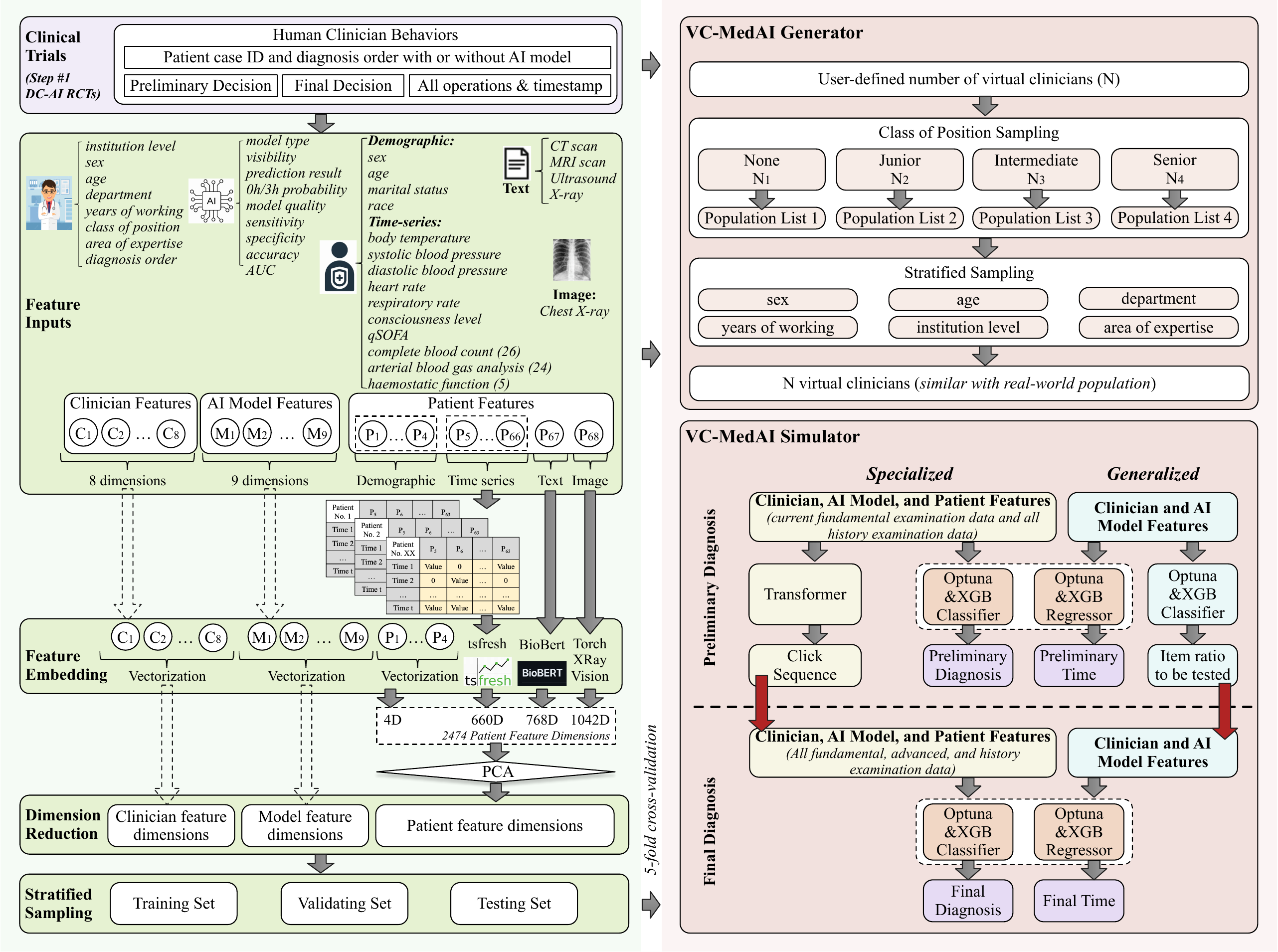}
\caption{\textbf{The Design and Implementation of VC-MedAI.} Based on Step \#1 DC-AI RCTs, 
VC-MedAI identifies a series of features in terms of clinician, AI model, and patient properties. 
The VC-MedAI generator generates user-defined number of virtual clinicians and reflect similar features with real-world human clinician population. 
VC-MedAI simulator receives feature input and simulates the operation behaviors, diagnosis decision, and time consumption during preliminary diagnosis stage, and outputs the final diagnosis and time consumption during final diagnosis stage.
}
\label{virtualdoctor}
\end{figure}

\noindent \textbf{The design and implementation of VC-MedAI clinician generator.} 
Given a user-defined number of virtual clinicians, the generator performs two-stage samplings according to real-world human clinician populations, as shown in Fig.~\ref{virtualdoctor}. 
First, to ensure the rationality of clinician features, e.g., avoiding mismatches in age, position, and experience, the generator samples the class of position and maintains a population list for each category, namely, none, junior, intermediate, and senior. Second, the generator performs stratified sampling and samples the other features like age according to population lists. 

\vspace{0.3in}

\noindent \textbf{The design and implementation of VC-MedAI behavior simulator.}
Fig.~\ref{virtualdoctor} presents the construction of VC-MedAI, including two-stage (preliminary and final) diagnosis simulations.
This aligns with real-world medical diagnosis, as clinicians typically rely on the fundamental items and history examination data to make a preliminary diagnose, while also requiring additional advanced examination items for a final decision.
We train VC-MedAI specialized simulators based on Step \#1 DC-AI RCTs as follows: 
(1) feature inputs from clinician, AI models, and patient perspectives. 
(2) feature embedding referring to the HAIM framework 
~\cite{soenksen2022integrated,christ2018time,lee2020biobert,cohen2022torchxrayvision}. 
(3) dimension reduction adopting principal components analysis (PCA)~\cite{mackiewicz1993principal} on patient features to prevent the clinician and AI model features from being overshadowed. After that, the number is 17, equal to the sum of clinician and AI model dimensions. 
(4) stratified sampling to divide a training, validating, and testing set with 7:2:1. 
(5) preliminary and final decision model training using 5-fold cross-validation. 
We use XGBClassifier~\cite{chen2016xgboost} and XGBRegressor~\cite{chen2016xgboost} for the simulation of diagnosis decisions and time, respectively. The hyperparameters are automatically optimized using Optuna~\cite{akiba2019optuna}. 
We use transformer model~\cite{vaswani2017attention} and XGBClassifier for the behavior simulation of click sequence and item ratio, respectively.
The current advanced examination data are excluded for the preliminary simulation while included for final one.
VC-MedAI generalized simulator follows similar processes but focuses solely on clinician and AI model features, excluding patient-related processes. 
Performance on testdata from Step \#1 DC-AI RCTs are in Appendix Fig. B1 and Tables B7-B12.


\backmatter





\bmhead{Acknowledgements}


We acknowledge supports from the Innovation Funding of ICT, CAS under Grant No. E461070.


\section*{Declarations}


\textbf{Competing interests}

\noindent The authors declare no conflicts of interest and no competing interests. \\

\noindent \textbf{Ethics statement}

\noindent This research has been approved by the Ethics Committee of Guilin Medical University (Approval No:GLMC20221101). All the participated clinicians have assigned the informed consent. The patient data used in this database are sourced from the publicly available MIMIC dataset, which the authors have received permission to use, ensuring that no new ethical issues are involved. \\

\noindent \textbf{Consent for publication}

\noindent All authors have read and approved the final manuscript. \\

\noindent \textbf{Data availability}

\noindent The diagnosis records collected from our two-step DC-AI RCTs have been organized into a series of CSV (comma-separated values) files and described in a companion paper about data descriptors~\cite{gao2024ai}. The database is publicly available from PhysioNet (waiting for approval) and International Open Benchmark Council (\url{https://www.benchcouncil.org/ai.vs.clinician/}). \\

\noindent \textbf{Code availability}

\noindent The VC-MedAI related code is publicly available from \url{https://github.com/BenchCouncil/VC-MedAI/}.  \\

\noindent \textbf{Author contribution}

\noindent W.G. conceptualized this study, formulated the designs, conceived the experiments, and wrote the manuscript. D.C., Z.Y., W.L., X,L., J.Z., J.X., and H.L. implemented the algorithms, collected and analyzed the data. L.M., N.Y., Y.K., D.L., P.P., W.H., Z.L., J.H., G.Z., C.J., F.H., T.W., and S.T. analyzed the data. B.X., Z.Z., Y.H., and J.Z. conceptualized this study, directed the project, and revised the manuscript. All authors have read and approved the final manuscript.

\bigskip






\bibliography{sn-bibliography}


\vspace{0.3in}

\newpage

\begin{appendices}

\section{Supplementary Methods}\label{secA1}


\textbf{10 patient-centered and clinician-centered (dual-centered) clinical settings.} 

\noindent The DC-AI RCTs are conducted across 14 medical centers collaborated with 125 human clinicians. The patient cases are in five groups. The 10 clinical settings are as follows: 

\begin{itemize}
    \item The patient cases of Group \#1 are diagnosed by human clinicians with and without the AI model assistance. Human clinicians can only see inference results of AI models while having no idea about the model characteristics like model name and model quality on testdata. 
    \begin{itemize}
        \item (1) diagnose without AI model assistance.
        \item (2) diagnose with a random model (control group with AUC 0.5).
        \item (3) diagnose with a LSTM-based low-quality (AUC 0.75) model.
        \item (4) diagnose with a LSTM-based medium-quality (AUC 0.85) model.
        \item (5) diagnose with a LSTM-based high-quality (AUC 0.95) model.
        \item (6) diagnose with a CoxPHM-based high-quality (AUC 0.95) model. \textbf{[Step \#2]}
    \end{itemize}

    \item For the other four patient groups, human clinicians make diagnosis decisions while being aware of the AI model's name and qualities.
    \begin{itemize}
        \item (7) The patient cases of Group \#2 are diagnosed by human clinicians with the assistance of a LSTM-based low-quality (AUC 0.75) model.
        \item (8) The patient cases of Group \#3 are diagnosed by human clinicians with the assistance of a LSTM-based medium-quality (AUC 0.85) model.
        \item (9) The patient cases of Group \#4 are diagnosed by human clinicians with the assistance of a LSTM-based high-quality (AUC 0.95) model.
        \item (10) The patient cases of Group \#5 are diagnosed by human clinicians with the assistance of a CoxPHM-based high-quality (AUC 0.95) model. \textbf{[Step \#2]}
    \end{itemize}
\end{itemize}

The patient cases within each experimental and control groups are randomly allocated to clinicians. Each clinician is allocated with six patient cases randomly from every clinical setting. The total diagnosis records is 7500 (125 clinicians * 6 patient cases * 10 settings).


\vspace{0.3in}

\noindent \textbf{The features used in VC-MedAI.}

\noindent (1) The clinician features contain eight dimensions. 
\begin{itemize}
    \item \emph{institution level} reflects the hospital ratings on the overall quality and performance. We have three rating categories including A, B, and C. 
    \item \emph{sex} indicates whether a clinician is male or female.
    \item \emph{age} indicates the age of a clinician.
    \item \emph{years of working} refers to the number of years a clinician has been employed or actively working in a hospital.
    \item \emph{department} refers to specific divisions or units within a hospital that focus on providing specialized medical care and services like emergency department. We have 14 department categories.
    \item \emph{class of position} represents the professional title of a clinician including chief physician, associate chief physician, attending physician, resident physician, junior physician.
    \item \emph{area of expertise} refers to a specific field or subfield where a clinician has acquired specialized skills and extensive experience.
    \item \emph{diagnosis order} indicates the clinician is diagnosing the $n^{th}$ patient in a particular setting like using or not using an AI model for assistance. We include this feature considering the order may influence psychological factors such as the clinician's trust in the AI model.
\end{itemize}

\noindent (2) The AI model features include nine dimensions.
\begin{itemize}
    \item \emph{model type} indicates the technique of the model such as  traditional machine learning and deep learning. We also introduce a random-based model type.
    \item \emph{model quality} defines the quality range of a model. Specifically, 1 indicates an AUC ranging from 0 to 0.6, while 2 signifies an AUC spanning 0.6 to 0.7. Moving forward, 3 represents an AUC falling between 0.7 and 0.8. Next, 4 denotes an AUC between 0.8 and 0.9, and finally, 5 represents an exceptional AUC ranging from 0.9 to 1.
    \item \emph{sensitivity} measures how well a model can identify patients with sepsis~\cite{swift2020sensitivity}.
    \item \emph{specificity} measures how well a model can identify people without sepsis~\cite{swift2020sensitivity}.
    \item \emph{accuracy} measures how well a model's predictions align with the actual or expected outcomes.
    \item \emph{AUC (Area Under the Curve)} provides a concise measure of the model's performance, with higher values indicating better classification ability.
    \item \emph{visibility} indicates whether the model characteristics are visible to the clinician, like the model type and quality.
    \item \emph{prediction result} indicates the diagnosis conclusion of a model. The prediction results are divided into three categories: sepsis, sepsis alert within 3 hours, and non-sepsis. Note that the 3-hour window aligns with the recommendations and guidelines on sepsis~\cite{henry2022factors,singer2016third}.
    \item \emph{0h/3h probability} depicts the probability of the sepsis onset at present (0h) and the sepsis onset within the next three hours (3h), respectively. Note that we use both 0h and 3h probability as features for VC-MedAI specialized simulator. For VC-MedAI generalized simulation, we provide separate simulators for 0h and 3h, considering that different diseases may involve different recommendations and guidelines on time windows. The data of generalized simulator illustrated in Results section derive from the VC-MedAI generalized simulator for 0h. The data from the VC-MedAI generalized simulator for 3h is supplemented in Appendix B, with Table B3-B6 showing the performance on a new AI model from Step \#2 DC-AI RCTs, Table B8-B12 showing the performance on testdata from Step \#1 DC-AI RCTs, and Table B13-B16 showing the performance on in-silico trials with 125 new virtual clinicians and 7500 virtual diagnosis records.
\end{itemize}

\noindent (3) The patient features consist of demographic information, time-series data, text, and images.
\begin{itemize}
    \item \emph{Demographic information} includes the sex, age, marital status, and race.
    \item \emph{Time-series examination items} include time-series fundamental items and advanced items. The included fundamental items are body temperature, systolic blood pressure, diastolic blood pressure, heart rate, respiratory rate, consciousness level, and qSOFA. The included advanced items are complete blood count (26 dimensions), arterial blood gas analysis (24 dimensions), haemostatic function (5 dimensions).
    \item \emph{Text data} includes the text notes of CT scans, MRI scans, ultrasound, and X-ray data.
    \item \emph{Image data} includes the chest X-ray images.
\end{itemize}

Feature embedding. For the time-series examination data, we construct a time-feature matrix and use tsfresh to generate a 660-dimensional embeddings for each patient. For the text data, we use BioBert to generate a 768-dimensional embeddings. For the image data, we use TorchXRayVision to generate a 1042-dimensional embeddings. After that, the clinician features have eight dimensions. AI model features have nine dimensions. The patient features have 2474 dimensions.

Data validation. Among 6000 diagnosis records from Step \#1 DC-AI RCTs, we filter the valid data for model training. For example, the records with the incomplete diagnosis or incomplete clinician information are removed. After the filtering, the total valid records are about 5500. In addition, we filter the diagnosis record with too short or too long diagnosis time, resulting in approximately 4800 records and 4900 records for preliminary and final diagnosis, respectively. 

\vspace{0.3in}

\end{appendices}

\end{document}